\documentclass[12pt,onecolumn]{article}

\usepackage[utf8]{inputenc} 
\usepackage[T1]{fontenc}    
\usepackage{hyperref}       
\usepackage{url}            
\usepackage{booktabs}       
\usepackage{amsfonts}       
\usepackage{nicefrac}       
\usepackage{microtype}      
\usepackage{lipsum}         
\usepackage{graphicx}       
\usepackage{amsmath}        
\usepackage{setspace}       
\usepackage{natbib}
\usepackage{multirow}

\usepackage{geometry}
\newcommand{\tucker}[1]{\left[\!\left[#1\right]\!\right]}
\newcommand{\bigsum}{\mathop{\vcenter{\hbox{\scalebox{2.0}{$\displaystyle\sum$}}}}}

\title{Convolution-Free Holistic Multivariance Decomposition Layer for Efficient Hyperspectral Image Classification Tensor Networks}

\author{
    Süha Tuna\textsuperscript{a,}\thanks{Corresponding author.}
    \;\;and Ülker Başar\textsuperscript{a,b}
    \\[2mm]
    \small
    \textsuperscript{a}Istanbul Technical University, Informatics Institute,\\
    \small
    Maslak 34469, \.Istanbul, Türkiye\\
    \small
    \textsuperscript{b}Istanbul Esenyurt University, Faculty of Business and Management Sciences,\\
    \small
    Department of Management Information Systems, Esenyurt, 34517, \.Istanbul, Türkiye
}

\date{}

\begin{document}

\maketitle

\begin{abstract}
Feature extraction for hyperspectral image classification is conventionally addressed using rigid tensor decompositions that fail to capture complex spatio-spectral interdependencies, or heavily parameterized convolutional neural networks that are computationally expensive. To overcome these limitations, this work introduces the Holistic Multivariance Decomposition (HMD) framework as a novel, end-to-end differentiable neural network layer. By explicitly separating independent single mode variations from cooperative higher dimensional interactions via learnable, matrix valued supports, the proposed HMD-0, HMD-1 and HMD-2 approximants are optimized jointly with a downstream classifier via backpropagation. Comprehensive evaluations across three benchmark HS datasets demonstrate that the higher level HMD layers achieve superior classification accuracy compared to classical learnable tensor baselines, including Tucker, Canonical Polyadic, and Tensor Train  decompositions. Furthermore, HMD-1 and HMD-2 achieve a generalization capacity and training stability comparable to standard 2D- and 3D-CNNs while requiring significantly fewer feature extractor parameters. These results demonstrate that the HMD framework provides a structurally robust substitute for traditional convolution in multidimensional HS image classification, offering high parameter efficiency and stability throughout the optimization process.
\end{abstract}


\section{Introduction}
\label{sec:intro}

Hyperspectral (HS) imaging represents a rapidly advancing frontier in remote sensing and data acquisition \citep{hsi_overview}. These complex data structures are typically captured via unmanned aerial vehicles (UAVs), drones, and satellites operating at various altitudes. By integrating rich informational content across both the spatial and spectral domains, hyperspectral imagery (HSI) embeds comprehensive information specific to a target geographic area \citep{hsi_overview,hs_apps_review}. An HS image consists of hundreds of narrow, contiguous spectral bands. These bands facilitate the precise differentiation of distinct objects within a region of interest. This high level of discrimination is possible because every material possesses a unique spectral signature. Because these detailed spectral signals provide fundamental information regarding underlying material components, they are widely exploited to resolve material classification problems from various fields. Consequently, the task of HS image classification has emerged as a crucial area that has consistently attracted significant scientific attention over the past decade \citep{hs_class}.

Beyond traditional remote sensing, HSI has found a wide variety of applications across diverse scientific disciplines. In agricultural and various industrial sectors, notable implementations of HSI include the
assessment of food quality and safety \citep{hs_food, hsi_agriculture}, as well as artwork authentication \citep{artwork} and the examination of 
counterfeit pharmaceuticals \citep{drug}. The technology is also gaining attraction in biomedical engineering \citep{corneal,gastric} and medical imaging \citep{medicalimaging} fields. Whether applied to environmental monitoring or clinical diagnostics, the dual presence of spatial and spectral data ensures robust pattern recognition \citep{pattern}, comprehensive image classification \citep{hs_class}, and precise spectral unmixing \citep{unmixing}.

To improve overall classification accuracy in HSI, a diverse array of feature extraction techniques has been developed and implemented \citep{pattern,hs_apps_review,hs_class}. Broadly speaking, these methodologies fall into two primary groups, conventional mathematical schemes and advanced learning based approaches. Within conventional frameworks, many data processing techniques originally designed for standard signals and images have been adapted for HSI. For instance, classical data reduction techniques like principal component analysis (PCA) \citep{pca}, linear discriminant analysis (LDA) \citep{lda} and factor analysis (FA) \citep{factor} are frequently applied along the spectral axis of HS images. These methods extract localized features from the corresponding spectral signals quite effectively. Although these techniques function well as feature extractors and yield satisfactory results, they fundamentally struggle to capture the complex spatial correlations inherent to HSI data.

To address this spatial limitation, researchers often apply standard feature extraction techniques designed for conventional color images \citep{feature}. These methods are run sequentially on each individual spectral band to elicit structural information from the HS cube. Within this sequential paradigm, transform based approaches such as the discrete wavelet transform (DWT) \citep{tuna_wavelet_hdmr} and the discrete cosine transform (DCT) \citep{dct} serve as vital analytical tools. However, despite their acknowledged success in localized feature extraction, these methods possess an inherent two dimensional nature. Consequently, they fail to simultaneously capture critical interband correlations across the spectral domain.

An alternative paradigm for feature extraction in HS imaging relies on tensor decomposition based techniques. Well known models such as the Tucker \citep{tucker_hsi} and Canonical Polyadic (CP) \citep{cp_hsi} decompositions have been widely applied to HS data. More recently, sequential core factorizations, such as the Tensor Train (TT) decomposition \citep{tt}, have also been explored for compact multidimensional representation. However, because these conventional methods depend on rigid low-rank approximations, they often fail to capture complex mode interrelations among distinct dimensions, particularly joint spatio–spectral features. To address this limitation, High Dimensional Model Representation (HDMR) \citep{tuna_wavelet_hdmr,efe_fusion,tuna_sparse_hdmr} and Enhanced Multivariance Products Representation (EMPR) \citep{tuna_tgrs,enis} frameworks have been effectively employed to isolate these spatio–spectral correlations. Unfortunately, both HDMR and EMPR lack flexible rank adjustment mechanisms. This inherently restricts the scope of the extracted information. To overcome these bottlenecks, a novel framework named Holistic Multivariance Decomposition (HMD) was recently proposed. HMD extracts highly efficient spatio–spectral features from HS images. Furthermore, it introduces a manageable dimensionality reduction parameter, allowing for dynamic adjustment of the retained subspace information.

In parallel with advances in tensor based feature extraction, learning based approaches have become a dominant paradigm for HS classification. Most notably, convolutional neural networks (CNNs) operate directly on raw spatio–spectral patches using two dimensional kernels that treat spectral bands as input channels or three dimensional kernels that convolve jointly over spatial and spectral axes. These models succeed owing to their capacity to learn discriminative features end-to-end, rather than relying on a fixed algebraic decomposition of each input patch.

To introduce learnability into the HMD framework, this work builds upon the foundational formulation by recasting the HMD-0, HMD-1, and HMD-2 approximants as fully differentiable feature extraction layers. Rather than remaining fixed a priori, their support matrices are optimized jointly with a shared downstream classifier via backpropagation. We benchmark these proposed layers against learnable Tucker, CP and TT layers of equivalent form, alongside conventional two dimensional and three dimensional CNN baselines. All methods are rigorously evaluated under a common training protocol, a shared classifier head, and a standard metric suite comprising Overall Accuracy, Average Accuracy, the macro-averaged F1-score, and Cohen's kappa coefficient.

Beyond classification accuracy, we place particular emphasis on parameter efficiency. Because every compared method shares an identical classifier head, the number of parameters within the feature extraction layer alone provides a direct, architecture independent  measure of representational cost. Consequently, the accuracy improvements delivered by each decomposition strategy can be evaluated directly against their exact parameter cost, an analysis not previously explored within the HMD framework. Finally, experiments are conducted across three benchmark HS datasets acquired by different sensors, spanning agricultural and urban land cover at varying spatial and spectral resolutions.

The remainder of this manuscript is organized as follows. Section \ref{sec:bg} reviews the mathematical foundations of tensor decomposition and multivariance representation frameworks relevant to HS feature extraction, together with the deep learning based spatio-spectral classification literature. Section \ref{sec:methods} details the proposed learnable HMD layers, the learnable Tucker, CP, and TT baselines, the 2D- and 3D-CNN baselines, and the unified classification architecture and training objective shared by all eight compared methods. Section \ref{sec:experiments} describes the benchmark datasets, preprocessing protocol, and training configuration. Section \ref{sec:results} presents and analyzes the comparative classification performance, parameter efficiency, training stability,
and cross dataset robustness of all eight methods. Finally, Section \ref{sec:conc} concludes the paper with a summary of key findings and directions for future research.

\section{Background}
\label{sec:bg}


In remote sensing and earth observation, HSI yields massive three dimensional data cubes. These structures are characterized by highly coupled spatial and spectral modes. Preserving these inherent geometric structures during feature extraction requires advanced
multilinear algebra frameworks such as tensor decomposition \citep{kolda,sidir,tensor_review}. Crucially, these frameworks
must remain capable of maintaining vital spatio-spectral correlations. Consequently, researchers rely on computationally efficient tensor decompositions to uncover latent patterns. The mathematical foundations
of this domain rest primarily on two classical methods. The first is the Tucker decomposition, which isolates the principal subspaces of the data by decomposing the tensor into a compact core tensor multiplied by factor matrices along each independent dimension \citep{tucker_hsi}. The second is the Canonical Polyadic (CP) decomposition \citep{cp_hsi}, which represents a higher order tensor as a minimal sum of rank-one vector outer products. A related but structurally distinct approach, the Tensor Train (TT) decomposition \citep{tt}, represents a tensor as a sequential chain of lower order core tensors contracted through shared bond indices, offering favorable scaling for higher order tensors at the cost of an inherently sequential factor structure rather than jointly multilinear.

A substantial portion of modern feature extraction relies on low-rank approximation phenomena. These techniques aim to reconstruct high dimensional tensors using a truncated summation of elementary components \citep{tensor_review}. In Tucker and CP frameworks, these components are structurally established by computing the outer product of individual vectors drawn from each independent dimension (mode). While this classical paradigm is mathematically elegant and computationally tractable, it suffers from a severe structural limitation. Since the underlying terms are rigidly tied to pure multilinear outer products, each rank-one component can only capture the isolated, independent variations of individual modes. Therefore, traditional low-rank approximations inherently overlook the complex, mutual relationships and highly coupled interdependencies that characterize the nature of HS data. When applied to HS datasets, this independence assumption forces the decomposition to sacrifice structural fidelity. Ultimately, this limitation prevents the framework from capturing the shared cross-mode dynamics, particularly the essential spatio-spectral correlations necessary for robust feature extraction \citep{kolda,sidir}.


To overcome the limitations associated with assuming strictly decoupled modal (e.g. spatial–spatial or spatio–spectral) behaviors, recent research has pivoted toward multidimensional analysis frameworks
explicitly engineered to map higher order interactions. Originating in function approximation, techniques such as HDMR decompose complex systems by systematically separating independent, single mode influences from their cooperative, higher dimensional interactions \citep{tuna_wavelet_hdmr,efe_fusion}. To enhance the representation capability of HDMR, the EMPR framework was introduced to provide a more descriptive representation of tensors. Unlike conventional tensor decompositions that rely strictly on rigid basis matrices, EMPR integrates flexible, one dimensional preselected support vectors directly into its structural representation \citep{tuna_tgrs,enis}.

Regardless of whether preselected or data-driven optimized support vectors are employed, the representation capability of EMPR remains inherently constrained \citep{tuna_tgrs}. Because these constituent support structures are strictly one dimensional, they can only capture information along a single direction at each mode. This structural limitation introduces a critical bottleneck in EMPR implementations. Overcoming this restriction which is transitioning from vector formed to matrix formed supports so that both spatial–spatial and spatio–spectral interactions within a given HS image are jointly preserved becomes precisely the motivation for the HMD framework. As formalized in the present work, HMD further distinguishes itself from HDMR and EMPR by ensuring that every isolated hierarchical component remains a full three dimensional tensor, and by exposing a single, controllable dimensionality reduction parameter $r$ that governs the trade-off between subspace compression and retained discriminative information.


Independently of the previously reviewed tensor algebraic frameworks, a parallel body of research has pursued spatio-spectral feature extraction through end-to-end trainable convolutional neural networks. These models learn discriminative filters directly from labeled patches rather than relying on a fixed algebraic decomposition. Early methodologies applied one dimensional convolutions exclusively along the spectral axis \citep{1dcnn}. Subsequent architectures extended this framework to two dimensional convolutions over the spatial extent of a patch, treating spectral bands as input channels \citep{2dcnn}. Furthermore, advanced models introduced three dimensional convolutions that operate jointly across both spatial axes and the spectral dimension within a single kernel \citep{3dcnn}. Patch based spatio-spectral classification, wherein a fixed size neighborhood around each labeled pixel is extracted and normalized prior to convolution, has consequently become a standard paradigm for HS image classification. This established framework represents the exact paradigm adopted for all eight feature extraction methods evaluated in this manuscript.

Relative to the tensor decomposition-based methods, 2D- and 3D-CNN feature extractors trade the explicit, interpretable low-rank structure of a tensor decomposition for the flexibility of an unconstrained, densely parameterized convolutional filter bank. This trade-off motivates the central empirical question addressed in this paper, which is whether the structured, low-rank feature extraction offered by HMD and its tensor decomposition counterparts (Tucker, CP, TT) can match or exceed the classification accuracy of conventional 2D- and 3D-CNN feature extractors at a
substantially lower parameter cost, when all methods are trained end-to-end under an identical protocol. 
This study formalizes all eight compared feature extraction layers in accordance with this established paradigm. Every layer operates as a differentiable mapping from an input HS patch to a fixed length feature vector. Consequently, this architectural choice enables the comparative evaluation to be executed within a unified framework.

\section{Methods}
\label{sec:methods}

An HS image patch is represented as a third order tensor $\mathcal{H} \in \mathbb{R}^{n_1 \times n_2 \times n_3}$, where the first two modes index the spatial extent of the patch and the third mode indexes the spectral bands. For a matrix $A \in \mathbb{R}^{n_i \times r_i}$, the mode-$n$ product of $\mathcal{H}$ with $A$, denoted $\mathcal{H} \times_n A$, contracts the $n$-th mode of $\mathcal{H}$ against the first index of $A$, so that $\mathcal{H} \times_n A$ tensor has size $n_1 \times \dots \times r_i \times \dots \times n_3$ \citep{kolda,sidir}. Following the literature, we further adopt the compact Tucker operator notation
\begin{equation}
\tucker{\mathcal{H} \, ; A, B, C} = \mathcal{H} \times_1 A \times_2 B \times_3 C,
\label{eq:tucker-operator}
\end{equation}
whenever the dimensions of $\mathcal{H}$ and of the mode matrices $A$, $B$, $C$ are compatible. Note that all eight feature extraction methods compared in this work are formulated as
learnable, differentiable mappings from an input patch 
$\mathcal{H}$ to a fixed length feature vector, trained end-to-end with an identical downstream classifier. This unified formulation ensures that comparisons of accuracy and parameter counts isolate the effects of the feature extractor structure, preventing confounding by incidental architectural differences.

\subsection{Tucker, CP and Tensor Train Decomposition Layers}
\label{subsec:baseline-tensor-layers}

Three established tensor decomposition strategies were implemented as learnable feature extraction layers. Each layer is parameterized by a set of trainable factor matrices. These matrices are optimized jointly with the classifier via backpropagation. The proposed  design choice remains consistent across all four tensor based methods evaluated in this study, including the proposed HMD framework. Consequently, this formulation allows their outputs to be interpreted as learned projections rather than isolated per-sample decompositions.

Given mode specific factor matrices $U_1 \in \mathbb{R}^{n_1 \times r_1}$, $U_2 \in \mathbb{R}^{n_2 \times r_2}$, $U_3 \in \mathbb{R}^{n_3 \times r_3}$, the Tucker feature layer computes the low-rank core
\begin{equation}
\mathcal{G} = \frac{1}{n_1 n_2 n_3} \tucker{\mathcal{H} \, ; U_1, U_2, U_3}
\in \mathbb{R}^{r_1 \times r_2 \times r_3},
\label{eq:tucker-core}
\end{equation}
followed by average pooling over the first two (spatial) modes to yield a $r_3$ dimensional feature vector. This is mathematically equivalent to the zeroth level HMD term defined in Eq.~\ref{eq:H0} and is included as a conventional baseline of identical mathematical form, allowing the incremental contribution of the higher degree HMD terms to be isolated
directly.

Given three factor matrices $A \in \mathbb{R}^{n_1 \times R}$, $B \in \mathbb{R}^{n_2 \times R}$, $C \in \mathbb{R}^{n_3 \times R}$ sharing a common rank $R$, the CP feature layer contracts the input patch against $R$ rank-one detector tensors formed from the columns of $A$, $B$, and $C$ as follows
\begin{equation}
\mathbf v_r = \frac{1}{n_1 n_2 n_3} \bigsum_{i,j,k} \mathcal{H}_{ijk} \, A_{ir} \,
B_{jr} \, C_{kr}, \qquad r = 1, \dots, R
\label{eq:cp-feature}
\end{equation}
yielding a feature vector $\mathbf v \in \mathbb{R}^{R}$. Unlike a complete CP reconstruction of the input tensor, Equation \ref{eq:cp-feature} is utilized exclusively as a feature extraction operator. Under this formulation, each resulting feature element quantifies the alignment of the HS patch with a specific rank-one spatio-spectral pattern. This mechanism functions analogously to applying a set of separable matched filters determined by the decomposition rank.

Given TT cores $G_1 \in \mathbb{R}^{n_1 \times R_1}$, $G_2 \in \mathbb{R}^{R_1 \times n_2 \times R_2}$, and $G_3 \in \mathbb{R}^{R_2 \times n_3 \times R_3}$, the TT feature layer performs the following sequential contraction
\begin{equation}
\mathcal{T} = \mathcal{H} \times_1 G_1, \qquad
\mathcal{T}' = \mathcal{T} \times_{(1,2)} G_2, \qquad
\mathbf v = \mathcal{T}' \times_{(1,2)} G_3,
\label{eq:tt-contraction}
\end{equation}
where $\times_{(1,2)}$ denotes simultaneous contraction over the bond index and the corresponding spatio-spectral mode, yielding a feature vector $\mathbf v \in \mathbb{R}^{R_3}$. This is the standard tensor train contraction scheme adapted here to act as a feature extraction operator rather than a full tensor approximation.

\subsection{Holistic Multivariance Decomposition}
\label{subsec:hmd}

Although Tucker and CP decompositions yield compact, single level tensor summaries and the TT approach models sequential mode dependencies, these representations lack the capacity to explicitly isolate the degree of spatio-spectral interaction. In particular, they do not differentiate between baseline zeroth order, independent single-mode first order, and interactive cross-mode second order variations within a given patch. To resolve this limitation, this work introduces the Holistic Multivariance Decomposition (HMD) framework. This approach fundamentally extends the High Dimensional Model Representation and Enhanced Multivariance Products Representation architectures. Rather than relying on one dimensional, vector valued support structures that restrict each factor to a single direction per mode, the proposed method introduces matrix valued supports capable of comprehensively capturing spatial-spatial and spatio-spectral interactions.

For each mode index, the HMD defines a support matrix with a reduced column dimension, represented as
\begin{equation}
  U_i \in \mathbb{R}^{n_i \times r_i}, \quad r_i < n_i, \quad i \in \{1, 2, 3\}
\end{equation}
which is constrained to be scaled semi-orthogonal according to the conditions  
\begin{equation}
\label{eq:hmd-support}
U_i^T\, U_i = n_i \, I_{r_i \times r_i}.
\end{equation}
Letting the complete support set be denoted by
\begin{equation}
\mathbf{U} = \{U_1, U_2, U_3\},
\end{equation}
while the subsets excluding the $i$-th and the $i$-th and $j$-th matrices are defined respectively as
\begin{equation}
\mathbf{U}^{(i)} = \{U_k \mid k \neq i\}, \qquad \mathbf{U}^{(i,j)} = \{U_k \mid k \neq i, j\}.
\end{equation}
Using the Tucker operator, the three dimensional HMD expansion decomposes the original input tensor into a hierarchy of multivariance components, expressed as 
follows
\begin{equation}
\label{eq:hmd3d}
\mathcal{H} = \tucker{\mathcal{H}_0; \mathbf{U}} + \sum_{i} \tucker{ \mathcal{H}_i; \mathbf{U}^{(i)} } + \sum_{i < j} \tucker{ \mathcal{H}_{i,j}; \mathbf{U}^{(i,j)} } + \mathcal{H}_{1,2,3}.
\end{equation}
In this hierarchical formulation, the zeroth degree component represents baseline spatio-spectral interaction, the three first degree components capture pure single-mode variation, the three second degree components capture pairwise cross-mode interactions, and the final term serves as the residual. Unlike HDMR and EMPR, every isolated HMD component remains a full three dimensional tensor, ensuring that spatial and spectral structures within each degree of interaction are preserved rather than collapsed.

Exploiting the commutativity of the mode product \citep{kolda,sidir} and the scaled semi-orthogonality identity in Eq. (\ref{eq:hmd-support}), the individual HMD components can be explicitly obtained by successive deflations. The zeroth degree term compresses the input tensor into a dense core that retains the global multidimensional structure of the patch, calculated as
\begin{equation}
\label{eq:H0}
\mathcal{H}_0 = \frac{1}{n_1 n_2 n_3} \tucker{ \mathcal{H}; U_1, U_2, U_3 } \in \mathbb{R}^{r_1 \times r_2 \times r_3}.
\end{equation}
The first degree terms isolate linear variation along a single mode after subtracting the corresponding zeroth degree projection, formulated as
\begin{equation}
\label{eq:Hi}
\begin{aligned}
\mathcal{H}_1 &= \frac{1}{n_2 n_3} \tucker{ \mathcal{H}; U_2, U_3 } - \tucker{ \mathcal{H}_0; U_1 } \in \mathbb{R}^{n_1 \times r_2 \times r_3}, \\
\mathcal{H}_2 &= \frac{1}{n_1 n_3} \tucker{ \mathcal{H}; U_1, U_3 } - \tucker{ \mathcal{H}_0; U_2 } \in \mathbb{R}^{r_1 \times n_2 \times r_3}, \\
\mathcal{H}_3 &= \frac{1}{n_1 n_2} \tucker{ \mathcal{H}; U_1, U_2 } - \tucker{ \mathcal{H}_0; U_3 } \in \mathbb{R}^{r_1 \times r_2 \times n_3}.
\end{aligned}
\end{equation}
The second degree terms isolate pairwise cross-mode interactions by further deflating the zeroth and first degree contributions, defined as
\begin{equation}
\label{eq:Hij}
\begin{aligned}
\mathcal{H}_{1,2} &= \frac{1}{n_3} \tucker{ \mathcal{H}; U_3 } - \tucker{ \mathcal{H}_0; U_1, U_2 } - \tucker{ \mathcal{H}_1; U_2 } - \tucker{ \mathcal{H}_2; U_1 } \in \mathbb{R}^{n_1 \times n_2 \times r_3}, \\
\mathcal{H}_{1,3} &= \frac{1}{n_2} \tucker{ \mathcal{H}; U_2 } - \tucker{ \mathcal{H}_0; U_1, U_3 } - \tucker{ \mathcal{H}_1; U_3 } - \tucker{ \mathcal{H}_3; U_1 } \in \mathbb{R}^{n_1 \times r_2 \times n_3}, \\
\mathcal{H}_{2,3} &= \frac{1}{n_1} \tucker{ \mathcal{H}; U_1 } - \tucker{ \mathcal{H}_0; U_2, U_3 } - \tucker{ \mathcal{H}_2; U_3 } - \tucker{ \mathcal{H}_3; U_2 } \in \mathbb{R}^{r_1 \times n_2 \times n_3}.
\end{aligned}
\end{equation}
With the index set defined across all three modes, the complete expansion admits a compact summation form, expressed as
\begin{equation}
\mathcal{H} = \bigsum_{I \subseteq \{1,2,3\}} \tucker{ \mathcal{H}_I; \mathbf{U}_I }.
\end{equation}
In practical applications, computing the full expansion is computationally expensive. Thus, truncating the expansion at a chosen degree yields the specialized approximants used as feature extractors in this study.

Three specific truncation levels are integrated as feature extraction layers. The zeroth level approximant, denoted as HMD-0, retains only the baseline term and is mathematically identical to the standard Tucker layer defined in Eq. (\ref{eq:tucker-core}). The first level approximant, denoted as HMD-1, retains the baseline term together with the three first degree terms. The second level approximant, denoted as HMD-2, retains the baseline term, all three first degree terms, and all three second degree terms (see Eq. (\ref{eq:hmd3d})). Because each component tensor beyond the zeroth degree baseline retains at least one unprojected mode of original dimension, a global average pooling operator $\varphi(\cdot)$ is applied over the two leading axes of every component to obtain a fixed length vector suitable for concatenation.

Given a component $\mathcal{H}_I$ of dimensions $n_{i_1}\times n_{i_2}\times n_{i_3}$, the operation $\varphi(\cdot)$ applied to $\mathcal{H}_I$ is explicitly formulated as follows
\begin{equation}
\varphi(\mathcal{H}_I) = \frac{1}{n_{i_1}n_{i_2}}\sum_{i_1} \sum_{i_2} \mathcal{H}_I\left(i_1, i_2, \cdot\right)
\end{equation}
The final layer output for a given truncation level is formed by concatenating the pooled components up to that degree, such that the feature vectors for HMD-0, HMD-1 and HMD-2 take the following explicit forms
\begin{equation}
\mathbf{v}^{(\text{HMD-0})} = \Big[\, \varphi(\mathcal{H}_0) \, \Big],
\end{equation}
\begin{equation}
\mathbf{v}^{(\text{HMD-1})} = \Big[\, \varphi(\mathcal{H}_0) \;\; \varphi(\mathcal{H}_1) \;\; \varphi(\mathcal{H}_2) \;\; \varphi(\mathcal{H}_3)\, \Big],
\end{equation}
and
\begin{equation}
\mathbf{v}^{(\text{HMD-2})} = \Big[\, \varphi(\mathcal{H}_0) \;\; \varphi(\mathcal{H}_1) \;\; \varphi(\mathcal{H}_2) \;\; \varphi(\mathcal{H}_3)\;\;\varphi(\mathcal{H}_{1,2})\;\;
\varphi(\mathcal{H}_{1,3})\;\;\varphi(\mathcal{H}_{2,3})\, \Big],
\end{equation}
respectively.

A structurally vital consequence of this formulation is that all three approximants share an identical set of trainable support matrices. Consequently, increasing the truncation level only alters which precomputed hierarchical components are pooled into the output feature vector, without increasing the number of learnable parameters in the feature extractor. This structural economy drives the parameter efficiency of HMD approximants, allowing higher order truncations to achieve superior accuracy at an identical parameter count. Because additional first and second degree terms derive from support matrices already required by the zeroth degree baseline rather than newly introduced weights, higher level approximants yield substantial performance gains without increasing the parameter footprint of the feature extractor.

The scaled semi-orthogonality condition in Eq. (\ref{eq:hmd-support}) is not imposed as a hard constraint during training, since the support matrices are learned end-to-end rather than obtained via a predetermined support generation process. Instead, semi-orthogonality is encouraged through a soft penalty added to the training loss for all three HMD approximants and the Tucker baseline, expressed as
\begin{equation}
\label{eq:L_ortho}
\mathcal{L}_{\text{ortho}} = \lambda \sum_{i=1}^{3} \left\lVert U_i^T\, U_i - n_i \, I_{r_i} \right\rVert_F^2,
\end{equation}
where the scaling parameter $\lambda$ represents a fixed regularization coefficient. An analogous Frobenius norm based weight penalty is applied to the TT cores and to the CP factor matrices scaled to their respective dimensions for those two baseline architectures, whereas no such penalty is required for the convolutional baselines.

\subsection{Convolutional Neural Network Baselines}
\label{subsec:cnn}

To situate the tensor decomposition layers relative to conventional spatial-spectral deep learning approaches, two convolutional feature extractors were implemented using minimal filter configurations. This restriction prevents either baseline's parameter count from trivially dominating the comparative experiments.

The 2D-CNN baseline treats the spectral dimension of the patch as input channels and applies two 2-D convolutional blocks of size $k\times k$ as 
\begin{equation}
X_1 = \sigma\Big(\mathrm{BN}\big(\mathrm{Conv2D}_{k \times k}(\mathcal{H})\big)\Big) ,
\qquad
X_2 = \sigma\Big(\mathrm{BN}\big(\mathrm{Conv2D}_{k \times k}(X_1)\big)\Big)
\label{eq:cnn2d}
\end{equation}
followed by global average pooling over the two spatial dimensions to yield the feature vector, where $\sigma$ denotes the ReLU activation and $\mathrm{BN}$ denotes batch normalization.

The 3D-CNN baseline instead reshapes the patch into a single-channel volume $\mathcal{H}' \in \mathbb{R}^{n_1 \times n_2 \times n_3 \times  1}$ and applies two 3-D convolutional blocks of size $k\times k\times k$,
\begin{equation}
X_1 = \sigma\Big(\mathrm{BN}\big(\mathrm{Conv3D}_{k \times k \times k}(\mathcal{H}')\big)\Big),
\qquad
X_2 = \sigma\Big(\mathrm{BN}\big(\mathrm{Conv3D}_{k \times k \times k}(X_1)\big)\Big)
\label{eq:cnn3d}
\end{equation}
followed by global average pooling over all three spatio-spectral dimensions. Unlike the 2D-CNN, this formulation convolves jointly over both spatial axes and the spectral axis, allowing local spatio-spectral correlations to be captured directly by the kernel rather than combined only at the channel aggregation step.

\subsection{Unified Classification Architecture and Training Objective}
\label{subsec:architecture}

All eight feature extraction methods described above are followed by an identical classifier head, so that classification performance and parameter count can be attributed to the feature extraction stage alone. The extracted feature vector $\mathbf v$ is passed through two fully connected blocks with batch normalization, ReLU activation, and dropout, followed by a final softmax layer as
\begin{equation}
\hat{y} = \operatorname{softmax} \Bigg\{ W_3 \cdot \operatorname{Dropout} \bigg[ \, \sigma \Big( \operatorname{BN} \big( W_2  \cdot \operatorname{Dropout} [ \, \sigma ( \operatorname{BN}(W_1 \cdot \mathbf v) ) \, ] \big) \Big) \, \bigg] \Bigg\},
\label{eq:classifier-head}
\end{equation}
with layer widths and dropout rate given in
Section~\ref{sec:experiments}. The complete network including feature extractor and the classifier head is trained end-to-end by minimizing the sparse categorical cross-entropy loss between $\hat{y}$ and the
ground truth label, plus the semi-orthogonality penalty in Eq.~(\ref{eq:L_ortho}) whose explicit definition is as follows
\begin{equation}
\mathcal{L} = \mathcal{L}_{\mathrm{CE}}\left(y, \hat{y}\right) + \mathcal{L}_{\text{ortho}}.
\label{eq:total-loss}
\end{equation}
Training hyperparameters (optimizer, learning rate, batch size, number of epochs, and the regularization coefficient $\lambda$) are held fixed across all eight methods and are reported together with the datasets and evaluation protocol in Section~\ref{sec:experiments}.

\section{Experimental Setup}
\label{sec:experiments}

\subsection{Datasets and Preprocessing}
\label{subsec:datasets}

The proposed feature extraction strategies were evaluated on three widely adopted benchmark HS datasets spanning agricultural, urban, and mixed land-cover settings which are Indian Pines, Pavia University and Loukia \citep{datasets}. Table \ref{tab:datasets} summarizes their properties.
\begin{table}[h!]
\centering
\caption{Summary of the HS datasets used in this study.}
\label{tab:datasets}
\begin{tabular}{llccc}
\hline
\textbf{Dataset} & \textbf{Sensor} & \textbf{Spatial Size} & \textbf{Spectral Bands} & \textbf{Classes} \\
\hline
Indian Pines & AVIRIS & $145 \times 145$ & 200 & 16                          \\
Pavia University & ROSIS           & $610 \times 340$           & 103                     & 9                           \\
Loukia           & Hyperion        & $250 \times 1376$          & 176                     & 14                          \\
\hline
\end{tabular}
\end{table}

The Indian Pines scene was acquired by the Airborne Visible/Infrared Imaging Spectrometer over an agricultural test site in northwestern Indiana, United States. It consists of $145 \times 145$ pixels at a spatial resolution of approximately 20~m, spanning the 0.4 to 2.5~$\mu\text{m}$ wavelength range across 200 retained spectral bands, and is annotated with 16 land-cover classes dominated by row-crop agriculture and forest cover. The Pavia University scene was acquired by the Reflective Optics System Imaging Spectrometer over the campus of the University of Pavia, Italy, at a spatial resolution of 1.3~m. It comprises $610 \times 340$ pixels across 103 spectral bands within the 0.43 to 0.86~$\mu\text{m}$ range and is labeled with nine urban land-cover classes, including asphalt, meadows, bitumen, and bricks. The Loukia scene, part of the HyRANK benchmark collected by the Hyperion sensor over Crete, Greece, provides a lower spatial resolution (30~m) and higher class count (14 classes across 176 spectral bands over a $250 \times 1376$ pixel extent) complement to the first two datasets. This dataset serves to assess the generality and robustness of each method beyond the conventional Indian Pines and Pavia University pairing that dominates prior tensor decomposition studies.

\begin{figure}[h!]
  \centering
  \includegraphics[width=.95\textwidth]{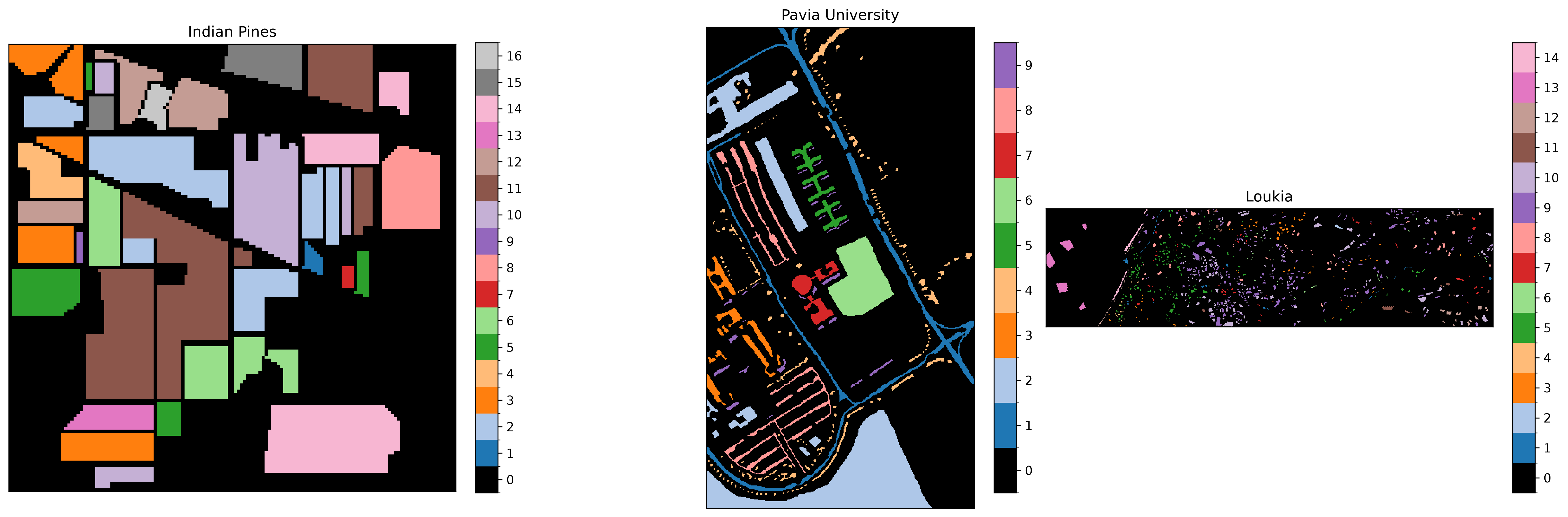}
  \caption{Ground truth classification maps for the three benchmark HS datasets, from left to right: Indian Pines, Pavia University and Loukia. Black pixels in all images represent unlabelled background regions excluded from training and evaluation.}
  \label{fig:gt}
\end{figure}

The corresponding ground truth images for the considered datasets are illustrated in Figure \ref{fig:gt}.

For each dataset, spectral bands were standardized to zero mean and unit variance per band across the full scene prior to spatial processing. The standardized cube was zero padded along its spatial dimensions to preserve border pixels, after which a spatio-spectral patch of size $11 \times 11 \times B$, where $B$ denotes the number of spectral bands, was extracted around every labeled pixel following the standard spatio-spectral patch based classification paradigm for HSI. Unlabeled background pixels were excluded from the sample set. Labeled samples were then partitioned into training and test sets using a stratified 80/20 split so that class proportions were strictly preserved across partitions. A fixed random seed  was applied across all partitioning and weight initialization operations to ensure that comparative evaluations across feature extraction methods were not confounded by sampling variance.


\subsection{Network Architecture and Training Configuration}
\label{subsec:network}

To isolate the effect of the feature extraction stage on classification performance, all compared methods share an identical downstream classifier head, differing solely in how the raw $11 \times 11 \times B$ patch is reduced to a compact feature vector. Eight feature extraction strategies were evaluated, spanning the three variants of the proposed HMD family, four established tensor decomposition baselines, and two convolutional baselines.

\begin{figure}[h!]
  \centering
  \includegraphics[width=.75\textwidth]{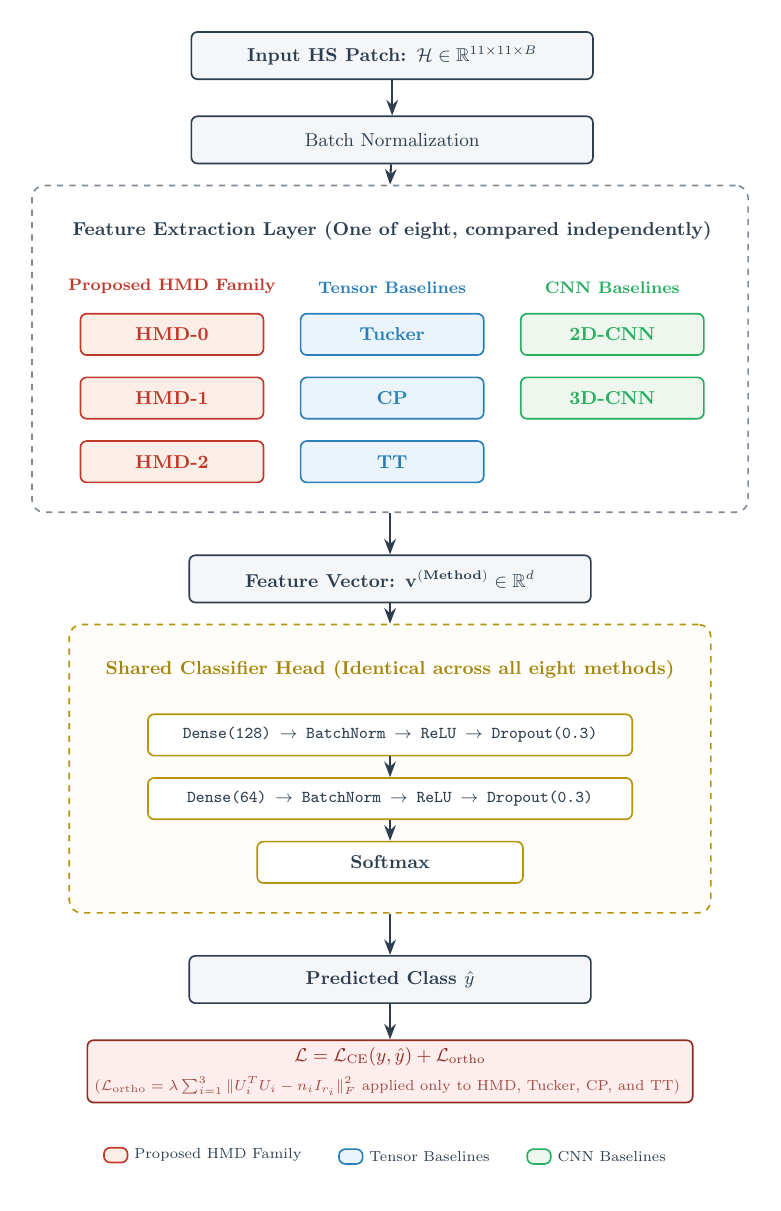}
  \caption{Schematic representation of the comparative deep learning pipeline for HS image classification. The architecture highlights the eight interchangeable feature extraction layers including the proposed HMD family, classical tensor models, and CNN baselines feeding into an identical, shared classifier head.}
  \label{fig:diagram}
\end{figure}

Within the proposed HMD family, the zeroth level HMD (HMD-0) performs a Tucker style multilinear projection of the patch onto a low-rank core with dimensions $(r_1, r_2, r_3)$, spatially pooled to a single descriptor. The first level approximant (HMD-1) extends HMD-0 by incorporating first order per-mode terms, analogous to the main effects of a functional analysis of variance decomposition of the input tensor. The second level approximant (HMD-2) further extends HMD-1 with second order pairwise-mode interaction terms, yielding the fullest holistic expansion among the three hierarchical variants.

The classical tensor baselines include the Tucker decomposition, representing a standard multilinear core projection included alongside the zeroth level HMD as a conventional baseline of equivalent mathematical form; the CP decomposition, which applies a single rank-$r$ polyadic factorization to the patch tensor; and the TT  decomposition, which performs a sequential three core contraction of the patch tensor.

To benchmark the tensor based layers against conventional spatio-spectral deep learning approaches, two convolutional feature extractors were implemented. The two dimensional convolutional neural network (2D-CNN) baseline employs a two layer architecture with sequential convolution, batch normalization, and activation blocks, followed by global average pooling that treats spectral bands as input channels. The three dimensional (3D-CNN) baseline applies a volumetric two layer architecture to the patch reshaped as a single channel volume, convolving jointly over spatial and spectral dimensions. This volumetric model evaluates whether modeling spatio-spectral locality via three dimensional kernels outperforms channel-wise spectral compression or low-rank decompositions.

Both convolutional baselines utilize minimal filter configurations to prevent their parameter counts from trivially dominating the evaluation. Specifically, the 2D-CNN employs 8 and 16 channels with $3\times 3$ kernels, while the 3D-CNN uses 4 and 8 channels with $3\times 3\times 3$ kernels across their respective layers. For all five tensor decomposition methods, a soft semi-orthogonality penalty on the mode-projection matrices or an analogous Frobenius norm weight penalty on the TT cores was incorporated into the training loss using a regularization coefficient $\lambda$. All decomposition ranks and convolutional filter counts remained fixed across datasets to strictly isolate the structural efficacy of the feature extractors.

Each feature extraction module is followed by an identical classifier head. The extracted feature vector passes through two fully connected blocks containing 128 and 64 units, respectively. Each block includes batch normalization, ReLU activation, and a $0.3$  dropout rate, culminating in a final softmax layer sized to the target classes. Input patches were batch normalized prior to feature extraction.

Models were trained using the Adam optimizer with an initial learning rate of $0.001$ for 50 epochs and a batch size of 32. The objective function minimized the sparse categorical cross-entropy loss, combined with the orthogonality penalty where applicable. Neither early stopping nor learning rate scheduling was utilized. Each method and dataset combination was evaluated independently using identical data splits and initialization seeds.

\subsection{Evaluation Metrics}
\label{subsec:metrics}

Model performance was assessed on the held-out test partition using four complementary metrics in accordance with standard practice in the HS image classification literature. Overall Accuracy (OA) measures the proportion of correctly classified test pixels across all classes. Average Accuracy (AA) calculates the unweighted mean of per-class recall, giving equal weight to minority and majority classes and thereby mitigating class imbalance bias. The F1-score (F1) computes the unweighted mean of per-class F1-scores to jointly capture precision and recall. Finally, Cohen's kappa ($\kappa$) coefficient provides a chance corrected measure of agreement between predicted and reference labels.

In addition to classification performance, the total number of trainable parameters was recorded for each configuration and explicitly decomposed into the parameters contained within the feature extraction module alone. This strategy allows the accuracy achieved by each HMD and baseline tensor layer to be interpreted relative to its parametric cost, enabling a direct efficiency comparison against the convolutional baselines. Because convolutional filters are inherently more parameter dense than the low-rank factor matrices utilized by tensor based methods, particularly for three dimensional kernels whose parameter count scales with the cube of the kernel size, both convolutional baselines were configured with the smallest filter counts that preserved a functioning two layer architecture, ensuring that the reported comparison reflects true accuracy-per-parameter efficiency.

All models were implemented in Python using the TensorFlow and Keras deep learning frameworks. Experiments were conducted under a Linux Ubuntu 24.04 LTS operating system equipped with an NVIDIA RTX 5060-Ti with 16 GB VRAM to ensure consistent computational benchmarking. To guarantee reproducibility, all NumPy and TensorFlow random seeds were fixed. The complete configuration for every experimental run was logged alongside per-epoch training curves and final test metrics.

\section{Results}
\label{sec:results}

\subsection{Overall Classification Performance}
\label{subsec:overall-performance}

Table \ref{tab:performance-indianpines}, Table \ref{tab:performance-paviau} and \ref{tab:performance-loukia} report the test set overall accuracy (OA), average accuracy (AA), macro F1 score (F1), and Cohen's kappa ($\kappa$) for all eight feature extraction strategies across the three datasets. The proposed HMD layers secured the top two positions across all metrics and scenes, with one exception on Pavia University. For this case, the 2D-CNN baseline achieved a marginally higher OA and AA than HMD-2 (99.88\% versus 99.31\% OA), albeit requiring over an order of magnitude more parameters. Specifically, HMD-1 delivered the best overall performance on Indian Pines with OA = 99.85\% and $\kappa$ = 0.998. 
\begin{table}[h!]
\centering
\caption{Test-set classification performance on Indian Pines. Best value in \textbf{bold}; second-best \underline{underlined}.}
\label{tab:performance-indianpines}
\begin{tabular}{lcccc}
\toprule
Method & OA (\%) & AA (\%) & F1 (macro) & $\kappa$ \\
\midrule
HMD-0 & 94.20 & 94.67 & 0.9421 & 0.9339 \\
\textbf{HMD-1} & \textbf{99.85} & \textbf{99.90} & \textbf{0.9975} & \textbf{0.9983} \\
\underline{HMD-2} & \underline{99.41} & \underline{99.64} & \underline{0.9935} & \underline{0.9933} \\
Tucker & 90.88 & 90.52 & 0.9110 & 0.8960 \\
CP & 74.93 & 64.56 & 0.6297 & 0.7159 \\
TT & 98.05 & 98.12 & 0.9825 & 0.9777 \\
2D-CNN & 94.00 & 96.43 & 0.9667 & 0.9319 \\
3D-CNN & 96.39 & 92.98 & 0.9288 & 0.9589 \\
\bottomrule
\end{tabular}
\end{table}

Furthermore, HMD-2 achieved the highest tensor based result on Pavia University with OA = 99.31\% with $\kappa$ = 0.991 alongside the top overall performance on Loukia with OA = 94.82\% and $\kappa$ = 0.938.
\begin{table}[h!]
\centering
\caption{Test-set classification performance on Pavia University. Best value in \textbf{bold}; second-best \underline{underlined}.}
\label{tab:performance-paviau}
\begin{tabular}{@{}lcccc@{}}
\toprule
Method & OA (\%) & AA (\%) & F1 (macro) & $\kappa$ \\
\midrule
HMD-0 & 78.87 & 70.78 & 0.7017 & 0.7229 \\
HMD-1 & 98.11 & 97.31 & 0.9691 & 0.9749 \\
\underline{HMD-2} & \underline{99.31} & \underline{98.91} & \underline{0.9882} & \underline{0.9909} \\
Tucker & 86.40 & 74.52 & 0.6981 & 0.8175 \\
CP & 89.61 & 82.16 & 0.8374 & 0.8590 \\
TT & 65.31 & 59.32 & 0.5191 & 0.5438 \\
\textbf{2D-CNN} & \textbf{99.88} & \textbf{99.74} & \textbf{0.9980} & \textbf{0.9985} \\
3D-CNN & 99.52 & 99.15 & 0.9904 & 0.9937 \\
\bottomrule
\end{tabular}
\end{table}

\begin{table}[h!]
\centering
\caption{Test-set classification performance on Loukia.
Best value in \textbf{bold}; second-best \underline{underlined}.}
\label{tab:performance-loukia}
\begin{tabular}{@{}lcccc@{}}
\toprule
Method & OA (\%) & AA (\%) & F1 (macro) & $\kappa$ \\
\midrule
HMD-0 & 79.34 & 70.77 & 0.7190 & 0.7554 \\
\underline{HMD-1} & \underline{91.23} & \underline{87.86} & \underline{0.8985} & \underline{0.8950} \\
\textbf{HMD-2} & \textbf{94.82} & \textbf{90.99} & \textbf{0.9292} & \textbf{0.9384} \\
Tucker & 73.27 & 71.33 & 0.6737 & 0.6823 \\
CP & 80.27 & 62.78 & 0.6490 & 0.7605 \\
TT & 78.71 & 71.94 & 0.7342 & 0.7409 \\
2D-CNN & 90.82 & 83.27 & 0.8698 & 0.8896 \\
3D-CNN & 77.82 & 69.60 & 0.6976 & 0.7365 \\
\bottomrule
\end{tabular}
\end{table}

The ultimate comparison among all feature extractors are provided in Table \ref{tab:mean-performance}.  Averaged across the three datasets, HMD-2 achieved the highest mean OA with $97.85\%$, AA with $96.51\%$, F1
with $0.9703$ and $\kappa$ with $0.9742$ of any method evaluated, followed by HMD-1. The 2D-CNN baseline ranked third on all four metrics, and the 3D-CNN baseline fourth. The remaining tensor decomposition baselines which are Tucker, CP and TT trailed the HMD family and the CNN baselines by a substantial margin on every metric, indicating that neither a bare
multilinear projection (Tucker) nor a single term factorization (CP) is sufficient to capture the spatio-spectral structure of these scenes. 
\begin{table}[h!]
\centering
\caption{Mean performance across the three datasets, ranked by mean
overall accuracy, alongside mean feature-extractor parameter count.}
\label{tab:mean-performance}
\begin{tabular}{@{}clccccc@{}}
\toprule
Rank & Method & OA (\%) & AA (\%) & F1 & $\kappa$ & Parameters \\
\midrule
1 & HMD-2   & 97.85 & 96.51 & 0.9703 & 0.9742 & 908 \\
2 & HMD-1   & 96.40 & 95.02 & 0.9550 & 0.9561 & 908 \\
3 & 2D-CNN  & 94.90 & 93.15 & 0.9449 & 0.9400 & 12{,}768 \\
4 & 3D-CNN  & 91.24 & 87.25 & 0.8723 & 0.8963 & 1{,}032 \\
5 & HMD-0   & 84.13 & 78.74 & 0.7876 & 0.8041 & 908 \\
6 & Tucker  & 83.51 & 78.79 & 0.7609 & 0.7986 & 908 \\
7 & CP      & 81.60 & 69.83 & 0.7054 & 0.7785 & 908 \\
8 & TT      & 80.69 & 76.46 & 0.7453 & 0.7541 & 4{,}322 \\
\bottomrule
\end{tabular}
\end{table}
Therefore, the first and second order interaction terms introduced by HMD-1 and HMD-2 account for the majority of the accuracy gain over the Tucker/HMD-0 baseline of equivalent zeroth order form.

\subsection{Subspace Dimensionality and Rank Sensitivity Analysis}
\label{subsec:ranks}

\begin{figure}[h!]
  \centering
  \includegraphics[width=\textwidth]{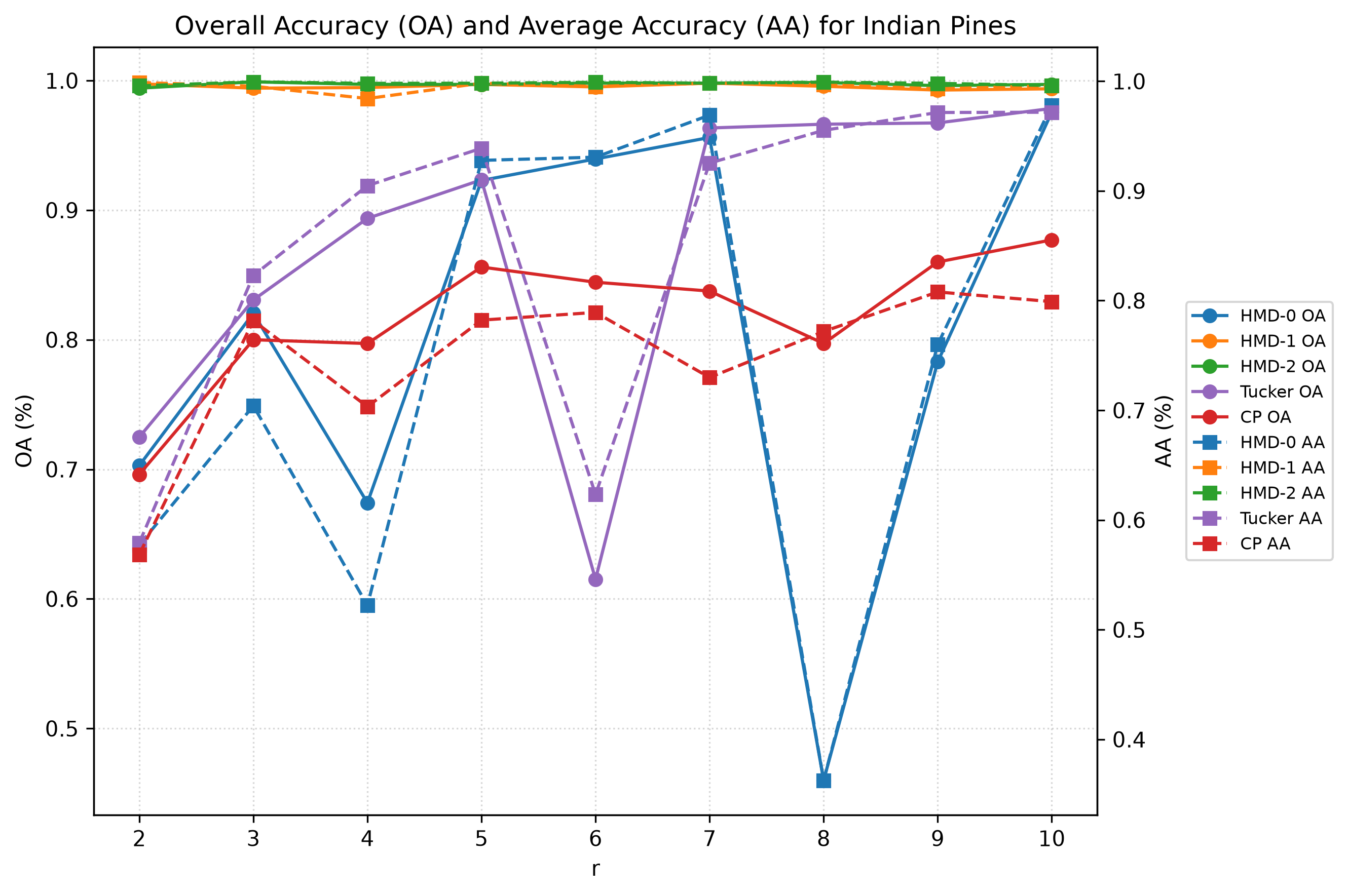}
  \caption{Sensitivity analysis of Overall Accuracy (OA) and Average Accuracy (AA) with respect to the subspace dimensionality reduction parameter ($r$) on the Indian Pines dataset.}
  \label{fig:rank_ip}
\end{figure}

\begin{figure}[h!]
  \centering
  \includegraphics[width=\textwidth]{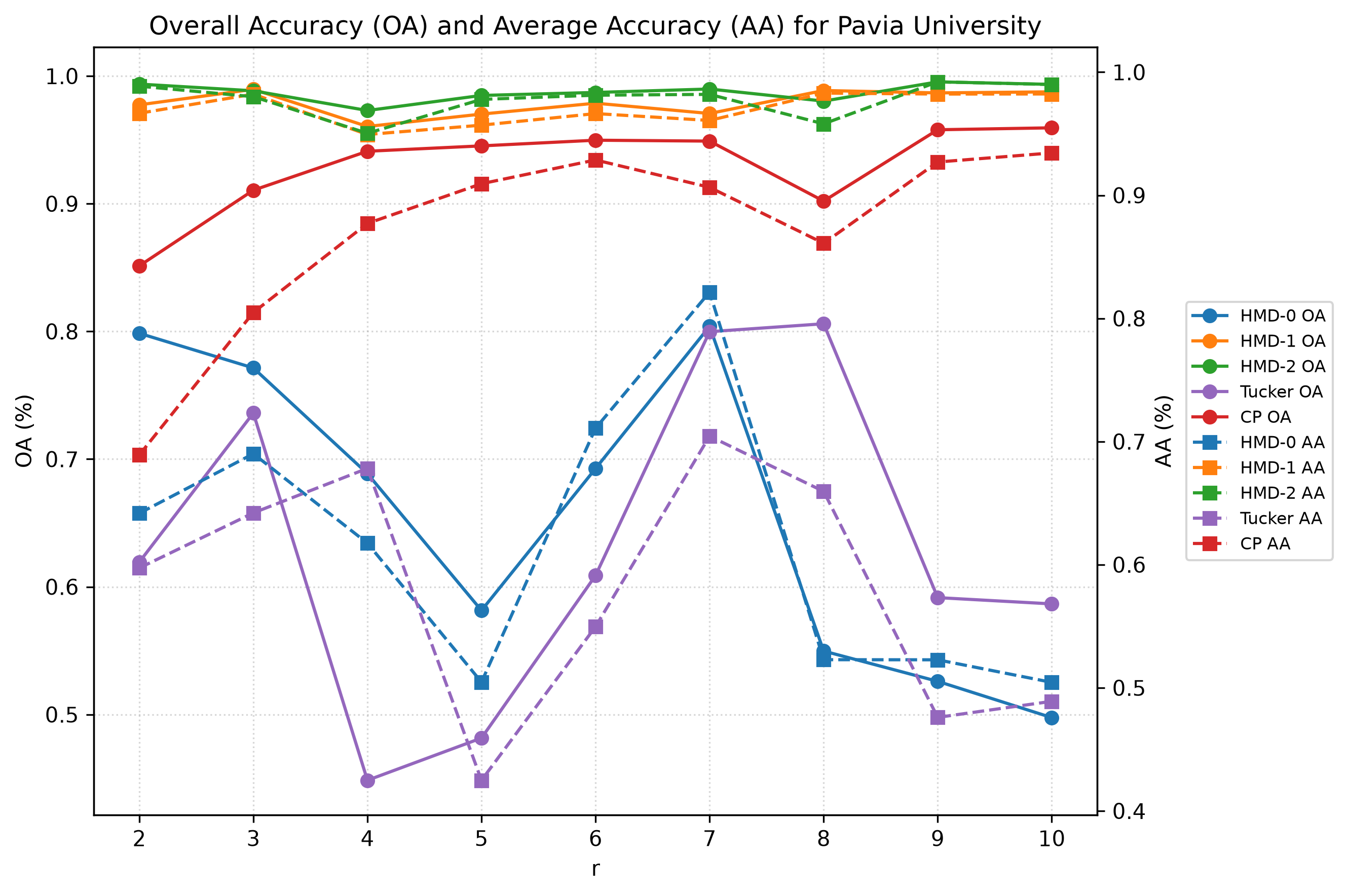}
  \caption{Sensitivity analysis of Overall Accuracy (OA) and Average Accuracy (AA) with respect to the subspace dimensionality reduction parameter ($r$) on the Pavia University dataset.}
  \label{fig:rank_pavia}
\end{figure}

\begin{figure}[h!]
  \centering
  \includegraphics[width=\textwidth]{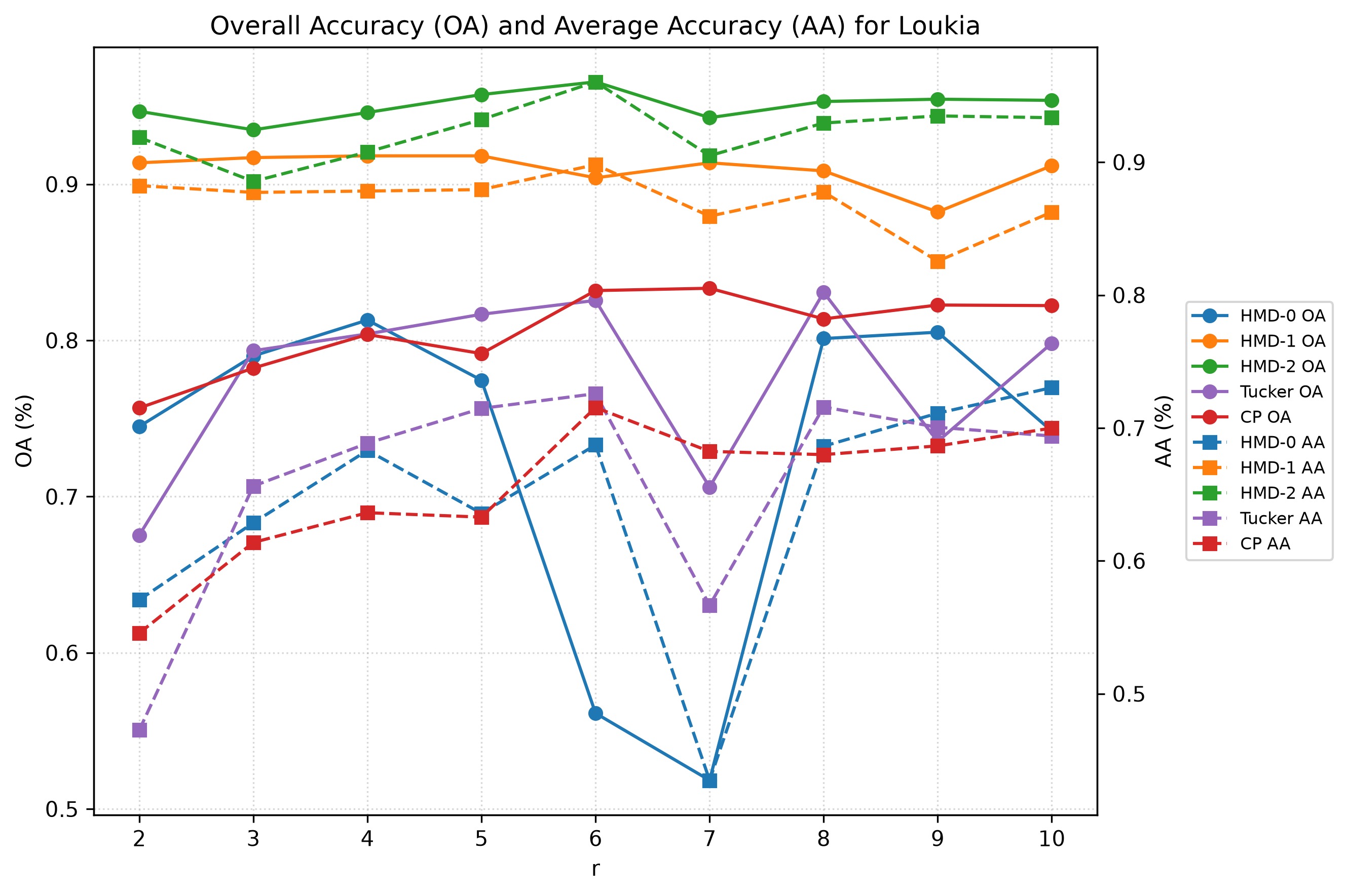}
  \caption{Sensitivity analysis of Overall Accuracy (OA) and Average Accuracy (AA) with respect to the subspace dimensionality reduction parameter ($r$) on the Loukia dataset.}
  \label{fig:rank_loukia}
\end{figure}

To rigorously evaluate the robustness of the extracted spatio-spectral features under varying levels of data compression, the classification performance of the proposed frameworks was analyzed across the subspace dimensionality reduction parameter $r \in [2, 10]$. The sensitivity of the OA and AA metrics for the Indian Pines, Pavia University, and Loukia datasets is illustrated in Figure \ref{fig:rank_ip}, Figure \ref{fig:rank_pavia} and Figure \ref{fig:rank_loukia}, respectively.

As depicted in the sensitivity curves for the Indian Pines dataset, the HMD-1 and HMD-2 establish overwhelming dominance and exceptional stability across the entire evaluated rank spectrum. While the classical Tucker decomposition and the zeroth level HMD-0 exhibit highly volatile behavior characterized by severe performance degradation at $r=4$ and a near total subspace collapse at $r=8$ where the higher level HMD frameworks maintain near optimal accuracy. This rapid convergence and sustained stability confirm that explicitly isolating linear directional multivariance components effectively shields the discriminative feature space from noise introduced by rank variations.

This behavioral disagreement is further amplified in the highly textured, high resolution urban geometries of the Pavia University scene. HMD-2 consistently achieves the highest classification accuracy, with HMD-1 providing a robust secondary baseline. The structural preservation of intricate, cross domain spatio-spectral interactions within HMD-2 proves highly resilient to parameter tuning. In contrast, the core tensor projections (HMD-0 and Tucker) display erratic oscillations, demonstrating that rigid low-rank approximations struggle to consistently encapsulate complex spatial dependencies. Furthermore, the CP  decomposition plateaus at a significantly lower accuracy ceiling, reaffirming its structural inability to map coupled multidimensional variations.

Finally, the evaluation on the Loukia HS scene substantiates the robust generalization of the proposed method. HMD-2 remains the superior framework, closely tracking with HMD-1 to yield highly stable, discriminative feature sets regardless of the chosen subspace dimension. Meanwhile, the HMD-0 and Tucker models once again exhibit tightly coupled, erratic performance trajectories, cratering notably around $r=7$.

Collectively, these results validate that the systematic extraction of higher degree multivariance terms successfully separates essential spatio-spectral features from latent noise. By overcoming the limitations of purely core dependent or strictly decoupled tensor approximations, the HMD framework ensures high fidelity HS classification even under volatile or tightly constrained subspace compressions.

\subsection{Training Stability and Convergence}
\label{subsec:training-stability}

To evaluate the learning behavior, optimization stability, and generalization capacity of the proposed frameworks, the epoch wise training and validation curves were analyzed across the benchmark datasets. The accuracy and loss trajectories over 50 training epochs are presented for the three datasets under consideration. 

Across all evaluated scenes, the heavily parameterized 2D-CNN baseline exhibits rapid training convergence, effectively memorizing spatio-spectral patterns due to its large parameter footprint. Among the tensor based architectures, the higher level HMD models, HMD-1 and HMD-2, demonstrate the strongest convergence, smoothly minimizing the training loss and closely tracking the CNN baseline's trajectory. Conversely, models relying on a rigid core tensor such as the classical Tucker decomposition, TT  and HMD-0 exhibit severe validation instability. This instability is characterized by abrupt drops in accuracy and massive loss spikes, indicating a deep vulnerability to subspace collapse during gradient descent.

\begin{figure}[h!]
  \centering
  \includegraphics[width=\textwidth]{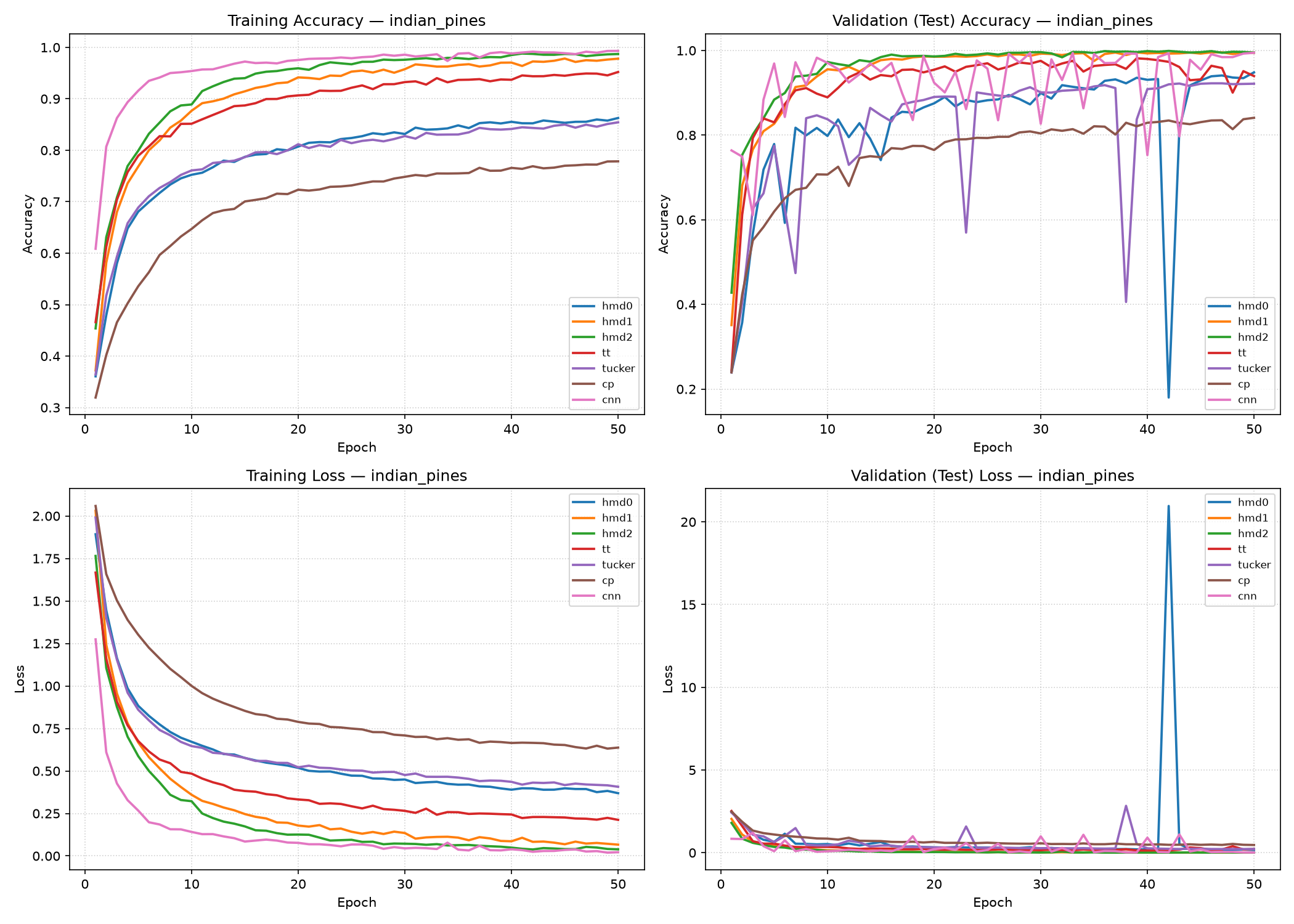}
  \caption{Training/validation accuracy and loss curves, all eight methods, Indian Pines dataset.}
  \label{fig:curves_ip}
\end{figure}
For the Indian Pines dataset, which is characterized by mixed agricultural parcels and high spectral similarity among vegetation classes, the training and validation curves illustrate the critical necessity of decoupling directional variations. While the CNN and higher level HMD models maintain stable validation trajectories that consistently exceed $90\%$ accuracy, the core dependent frameworks (Tucker, HMD-0, and TT) experience catastrophic validation drops, occasionally decreasing below $50\%$. This extreme volatility suggests that forcing highly mixed agricultural spectral signatures through a single dense core tensor exacerbates gradient perturbations during backpropagation. By explicitly isolating spatial and spectral fibers using $\mathcal{H}_1, \mathcal{H}_2, \mathcal{H}_3$, HMD-1 effectively insulates the decision boundaries from this latent noise, yielding the smooth and robust generalization necessary for agricultural classification.

The learning dynamics on the Pavia University scene further emphasize the structural advantages of the proposed framework when processing high resolution urban geometries. This dataset contains highly textured structures, localized shadows, and intricate spatial borders. The validation curves reveal that while the CP  decomposition converges smoothly, it fundamentally plateaus at a low accuracy ceiling due to its structural inability to map coupled multidimensional variations. Meanwhile, Tucker, TT, and HMD-0 suffer from repeated, drastic destabilization. HMD-2, however, successfully mirrors the stable generalization of the heavily parameterized CNN. By explicitly modeling pairwise cross-domain interactions ($\mathcal{H}_{1,2}, \mathcal{H}_{1,3}, \mathcal{H}_{2,3}$), HMD-2 directly resolves the complex spatio-spectral dependencies inherent to urban materials, providing a highly stable optimization landscape that does not collapse under fine grained spatial texture variations.

\begin{figure}[h!]
  \centering
  \includegraphics[width=\textwidth]{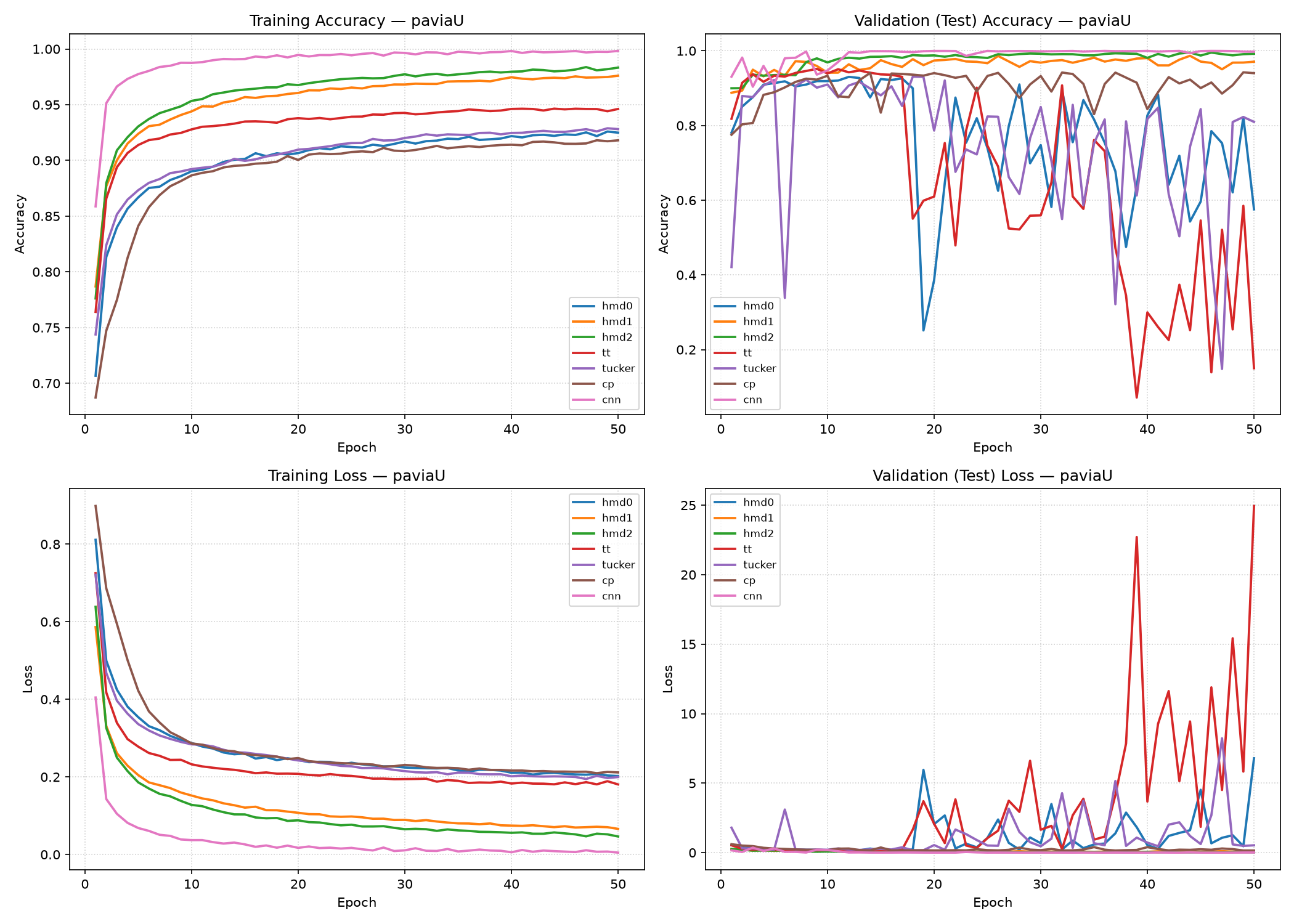}
  \caption{Training/validation accuracy and loss curves, all eight methods, Pavia University dataset.}
  \label{fig:curves_pu}
\end{figure}
On the Loukia dataset, characterized by complex vegetation and geological distributions, validation loss curves distinctly highlight the structural fragility of sequential and bottleneck tensor approximations. Specifically, the TT and Tucker models exhibit sudden, significant spikes in validation loss alongside corresponding accuracy degradation. The first- and second-level HMD frameworks bypass this volatility entirely. They maintain tightly bounded, low variance validation trajectories that parallel the convolutional neural network baseline. By distributing the representational load across isolated multivariance terms rather than imposing an artificial sequential order or a singular bottleneck, the proposed approach successfully captures the underlying signatures while avoiding optimization instability.

\begin{figure}[h!]
  \centering
  \includegraphics[width=\textwidth]{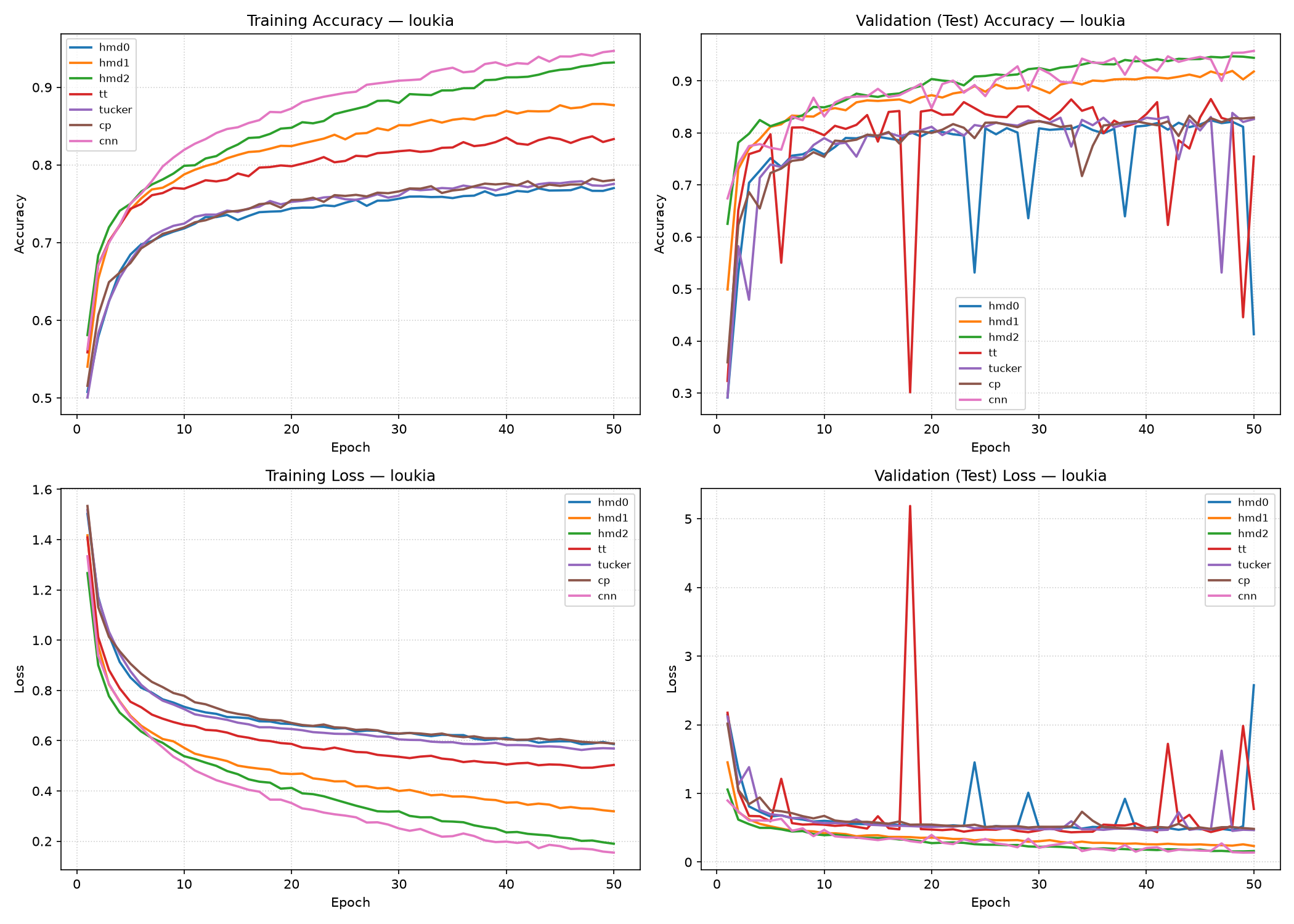}
  \caption{Training/validation accuracy and loss curves, all eight methods, Loukia dataset.}
  \label{fig:curves_loukia}
\end{figure}
Ultimately, these empirical optimization results validate the theoretical superiority of the HMD  expansion. While traditional tensor layers force all optimization pressure onto a rigid core tensor resulting in unstable decision boundaries and erratic validation behavior the higher level HMD layers systematically distribute this load. Consequently, HMD-1 and HMD-2 achieve a highly stable optimization landscape that rivals the generalization capacity of standard convolutional networks, while utilizing only a fraction of the computational parameters.

The relative performance rankings of the evaluated methods remained largely consistent across the three datasets, with HMD-1 and HMD-2 consistently securing the highest tiers of classification accuracy. However, the performance margins over the established baselines varied significantly depending on specific scene characteristics. The disparity between the proposed higher level HMD frameworks and the classical tensor baselines (Tucker, CP, HMD-0) was most pronounced on the Pavia University and Loukia datasets, yielding improvements of up to $20.4$ and $21.6$ OA percentage points between HMD-2 and HMD-0, respectively. Conversely, this gap narrowed on the Indian Pines scene, where even the least effective tensor model which is CP method with OA = 74.93\%, remained competitive with the strongest non-HMD baseline.

Among all evaluated architectures, the relative standing of the 2D-CNN baseline exhibited the highest degree of scene dependency. While it emerged as the highest performing method on Pavia University, it was outperformed by HMD-1 and HMD-2 by $3.6$ and $6.7$ OA points, respectively, on the Loukia scene. Because Loukia is characterized by the fewest labeled samples per class among the evaluated datasets, this performance degradation indicates that the substantial parametric footprint of standard convolutional networks becomes a computational liability rather than an asset under constrained training data conditions. 

\subsection{Parameter Efficiency}
\label{subsec:parameter-efficiency}

Every feature extraction layer evaluated in this work is defined by an explicit set of weight tensors, allowing for a closed-form parameter count determined strictly by the patch dimensions ($n_1$, $n_2$, $n_3$), the chosen rank or filter configuration, and the kernel size for the convolutional baselines. Deriving these mathematical forms directly from the HMD and CNN formulations clarifies how the representational cost of each method scales with the number of spectral bands, $n_3$. Because $n_3$ varies significantly across the three benchmark datasets, these theoretical formulations provide clear predictions that can be directly verified against the empirical parameter counts.

For the HMD-0, HMD-1, HMD-2 and Tucker layers, the only trainable variables are the support matrices $U_1 \in \mathbb{R}^{n_1 \times r}$, $U_2 \in \mathbb{R}^{n_2 \times r}$, and $U_3 \in \mathbb{R}^{n_3 \times r}$ introduced in Eq. \ref{eq:hmd-support}. These layers do not require additional internal bias or normalization parameters. As a result, all four models share an identical parameter count, that is
\begin{equation}
P^{(\text{HMD-0,1,2})} = P^{(\text{Tucker})} = r(n_1 + n_2 + n_3)
\end{equation}
This equivalence is a direct arithmetic consequence of the structural properties established in Section \ref{subsec:hmd} increasing the HMD truncation level only alters how the computed hierarchical components are pooled, without introducing new learnable weights. Similarly, the CP layer relies on three factor matrices with shapes identical to $U_1, U_2$, and $U_3$. Thus, for equivalent rank configurations, the CP parameter count is computed as follows
\begin{equation}
 P^{(\text{CP})} = R(n_1 + n_2 + n_3)
\end{equation}
which is numerically indistinguishable from $P^{(\text{HMD-0,1,2})}$. Therefore, any performance disparities between the CP model and the HMD family  reflect differences in representational architecture rather than parameter capacity. 

In contrast, the TT layer scales quadratically with respect to its rank. Its intermediate core tensor, $G_2 \in \mathbb{R}^{R_1 \times n_2 \times R_2}$, simultaneously carries two bond dimensions. Assuming $R_1 = R_2 = R_3 = r$ and spatial dimensions $n_1 = n_2 = 11$, the TT parameter count becomes 
\begin{equation}
P^{(\text{TT})} = 11r + r^2(11 + n_3)
\end{equation}
This quadratic scaling causes the TT layer to grow significantly faster with $r$ than the HMD, Tucker and CP frameworks.

The convolutional baselines require a different parameter accounting structure. Each convolutional block introduces kernel weights, a bias vector, and batch normalization parameters. Specifically, for a Conv2D or Conv3D block featuring $C_{\text{in}}$ input channels and $f$ output filters, the kernel accounts for $k^d\, C_{\text{in}}\, f$ weights, where $d \in \{2, 3\}$ denotes the kernel dimensionality. The bias vector adds additional $f$ parameters, and the batch normalization layer contributes $4f$ parameters comprising two trainable parameters, $\gamma$ and $\beta$, alongside two non-trainable parameters for the running mean and variance. All these components are included in the overall parameter counts, resulting in $k^d\,C_{\text{in}}\, f + 5f$ parameters per block. In the 2D-CNN baseline, the spectral dimension defines the input channel count of the initial convolutional block. Consequently, its total parameter count is computed as follows  
\begin{equation}
P^{(\text{2D-CNN})} = k^2 f_1(n_3 + f_2) + 5(f_1 + f_2)
\end{equation}
where $f_1$ and $f_2$ denote the filter counts for the first and second convolutional layers, respectively. Evidently, $P^{(\text{2D-CNN})}$ scales linearly with $n_3$, mirroring the scaling behavior of the tensor-based layers.

Conversely, the 3D-CNN operates on a patch reshaped into a single channel volume prior to convolution. Here, the spectral axis is absorbed directly into the kernel's volumetric receptive field rather than serving as the channel count. As a result, its parameter equation becomes
\begin{equation}
P^{(\text{3D-CNN})} = k^3 f_1(1 + f_2) + 5(f_1 + f_2)
\end{equation}
that is entirely independent of $n_3$. This invariance to the spectral band count is a unique structural characteristic not shared by any of the other evaluated models.
\begin{table}[h!]
\centering
\caption{Closed-form feature-extractor parameter counts by method and
dataset, at the rank/filter configuration used throughout this study
($r=5$; TT: $R_1=R_2=R_3=5$; 2D-CNN: $k=3$, $f_1=8$, $f_2=16$; 3D-CNN:
$k=3$, $f_1=4$, $f_2=8$). Indian Pines: $n_3=200$; Pavia University:
$n_3=103$; Loukia: $n_3=176$.}
\label{tab:param-complexity}
\begin{tabular}{@{}llccc@{}}
\toprule
Method & Formula & Indian Pines & Pavia University & Loukia \\
\midrule
HMD-0\textsuperscript{*} & $r(n_1+n_2+n_3)$ & 1{,}110 & 625 & 990 \\
HMD-1\textsuperscript{*} & $r(n_1+n_2+n_3)$ & 1{,}110 & 625 & 990 \\
HMD-2\textsuperscript{*} & $r(n_1+n_2+n_3)$ & 1{,}110 & 625 & 990 \\
Tucker\textsuperscript{*} & $r(n_1+n_2+n_3)$ & 1{,}110 & 625 & 990 \\
CP\textsuperscript{*} & $R(n_1+n_2+n_3)$ & 1{,}110 & 625 & 990 \\
TT & $n_1R_1+R_1n_2R_2+R_2n_3R_3$ & 5{,}330 & 2{,}905 & 4{,}730 \\
2D-CNN & $k^2f_1(n_3+f_2)+5(f_1+f_2)$ & 15{,}672 & 8{,}688 & 13{,}944 \\
3D-CNN & $k^3f_1(1+f_2)+5(f_1+f_2)$ & 1{,}032 & 1{,}032 & 1{,}032 \\
\bottomrule
\end{tabular}
\end{table}

Table \ref{tab:param-complexity} provides these closed form mathematical equations for the three benchmark datasets. The calculations utilize the standardized rank and filter configurations held constant throughout the experiments: $r = 5$ for the HMD family, Tucker, and CP models; $R_1 = R_2 = R_3 = 5$ for TT; $k = 3$, $f_1 = 8$, and $f_2 = 16$ for the 2D-CNN; and $k = 3$, $f_1 = 4$, and $f_2 = 8$ for the 3D-CNN. Because the Indian Pines, Pavia University, and Loukia datasets span a broad spectrum of spectral bands containing $n_3 = 200$, $103$, and $176$, respectively, the parameter counts for all $n_3$ dependent methods adjust proportionately. Loukia correctly occupies an intermediate parametric position between Pavia University and Indian Pines for these methods. Meanwhile, the structural design of the 3D-CNN ensures its parameter count remains rigidly fixed at $1032$ across all datasets, regardless of spectral resolution.

\section{Conclusion}
\label{sec:conc}

This study successfully implements the Holistic Multivariance Decomposition framework as a series of end-to-end trainable, convolution-free feature extraction layers for HS image classification. By systematically isolating linear directional components and pairwise cross-domain interactions, the proposed HMD-1 and HMD-2 layers address the structural bottlenecks of classical tensor decompositions such as the Tucker, Canonical Polyadic, and Tensor Train models which rely on rigid, low-rank core projections and suffer from optimization volatility and subspace collapse. Empirical evaluations across diverse agricultural, urban, and mixed land-cover scenes demonstrate that HMD-2 and HMD-1 consistently yield state-of-the-art classification performance, achieving mean Overall Accuracies of 97.85\% and 96.40\%, respectively. Crucially, this discriminative advantage is realized with exceptional parameter efficiency. The proposed methods match the minimal footprint of baseline core tensor models while requiring substantially fewer trainable weights than conventional CNN architectures.

Collectively, these findings demonstrate that the discriminative advantage of the HMD family is not restricted to a specific sensor modality or class distribution. Rather, its structural superiority is most evident in challenging regimes characterized by limited labeled samples and high class counts representing a scenario of paramount practical relevance for operational hyperspectral mapping.

The framework exhibits remarkable structural robustness, insulating the latent feature space from optimization noise even under highly constrained rank configurations or limited training regimes. Ultimately, the Holistic Multivariance Decomposition provides a computationally lightweight, stable, and interpretable deep learning layer, presenting a highly efficient alternative to standard convolutional networks for advanced hyperspectral classification.

\bibliographystyle{plainnat}
\bibliography{refs}

\end{document}